# IDENTIFYING LEXICAL PARAPHRASES FROM A SINGLE CORPUS: A CASE STUDY FOR VERBS

**Oren Glickman and Ido Dagan**

Computer Science Department

Bar Ilan University

Ramat Gan 52900, Israel

{glikmao,dagan}@cs.biu.ac.il

## Abstract

This paper studies the potential of identifying lexical paraphrases within a single corpus, focusing on the extraction of verb paraphrases. Most previous approaches detect individual paraphrase instances within a pair (or set) of "comparable" corpora, each of them containing roughly the same information, and rely on the substantial level of correspondence of such corpora. We present a novel method that successfully detects isolated paraphrase instances within a single corpus without relying on any a-priori structure and information. A comparison suggests that an instance-based approach may be combined with a vector-based approach in order to assess better the paraphrase likelihood for many verb pairs.

## 1 Introduction

The importance of paraphrases has been recently receiving growing attention. Broadly speaking, paraphrases capture core aspects of variability in language, by representing (possibly partial) equivalencies between different expressions that correspond to the same meaning. Representing and tracking language variability is critical for many applications (Jacquemin 99). For example, a question might use certain words and expressions while the answer, to be found in a corpus, might include paraphrases of the same expressions (Hermjakob et al. 02). Another example is multi-document summarization (Barzilay et al. 99). In this case, the system has to deduce that different expressions found in several documents express the same meaning; hence only one of them should be included in the final summary.

Recently, several works addressed the task of acquiring paraphrases (semi-) automatically from corpora. Most attempts were based on identifying corresponding sentences in parallel or "comparable" corpora, where each corpus is known to include texts that largely correspond to texts in another corpus (see next section). The major types of comparable corpora are different translations of the same text, and multiple news sources that overlap largely in the stories that they cover. Typically, such methods first identify pairs (or sets) of larger contexts that correspond to each other, such as corresponding documents, by using clustering or similarity measures at the document level, and by utilizing external information such as requiring that corresponding documents will be from the same date. Then, within the corresponding contexts, the algorithm detects individual pairs (or sets) of sentences that largely overlap in their content and are thus assumed to describe the same fact or event.

| 20-08-1996 | 16-09-1996 |
|---|---|
| …The broadcast, which gave no source for the information, followed a flurry of rumours that Lien had arrived in various European nations. China regards Nationalist-ruled Taiwan as a rebel province ineligible for foreign ties and has sought to isolate it diplomatically **since a civil war separated them in 1949**. Adomaitis said Ukraine maintains only economic relations with Taiwan with no political or diplomatic ties… | "I recognise there are political issues, but I nevertheless see it as a golden opportunity for Taiwan to increase its role in this important international organisation, and to play the part that it should as a major Asian economy," Summers said. China, which has regarded Taiwan as a rebel province **since a civil war split them in 1949**, says the island is not entitled to membership as a sovereign nation in international bodies. Beijing has said it would accept Taiwan's membership in the WTO as a customs territory, but not before China itself is allowed to join the world trade club. |

Table 1: example of extracting the lexical paraphrase <separate, split> from distinct stories

(Lin & Pantel 01) propose a different approach for extracting "inference rules", which largely correspond to paraphrase patterns. Their method extracts such paraphrases from a single corpus rather than from a comparable set of corpora. It is based on vector-based similarity, which compares typical contexts in a "global" manner rather then identifying all actual paraphrase instances that describe the same fact or event.

The goal of our research is to explore further the potential of learning paraphrases within a single corpus. Clearly, requiring a pair (or set) of comparable corpora is a disadvantage, since such corpora do not exist for all domains, and are substantially harder to assemble. On the other hand, the approach of detecting actual paraphrase instances, as was previously achieved within comparable corpora, seems to have high potential for extracting reliable paraphrase patterns. We therefore developed a method that detects concrete paraphrase instances within a single corpus. Such paraphrase instances can be found since a coherent domain corpus is likely to include repeated references to the same concrete facts or events, even though they might be found within generally different "stories" (see Table 1 for a paraphrase example extracted by our system originating from distinct stories).

The first version of our algorithm was restricted to identify lexical paraphrases of verbs, in order to study whether the approach as a whole is at all feasible. The challenge addressed by our algorithm is to identify isolated paraphrase instances that describe the *same* fact within a single corpus. Such paraphrase instances need to be distinguished from instances of *distinct* facts that are described in similar terms. These goals are achieved through a combination of statistical and linguistic filters and a probabilistically motivated paraphrase likelihood measure. We found that the algorithmic computation needed for detecting such local paraphrase instances across a single corpus should be quite different than previous methods developed for comparable corpora, which largely relied on a-priori knowledge about the correspondence between the different stories from which the paraphrase instances are extracted.

We have further compared our method to the vector-based approach of (Lin & Pantel 01), which measures global similarity across all instances. The precision of the two methods on common verbs was comparable, but they exhibit some different behaviors. In particular, our instance-based approach seems to help assessing the reliability of candidate paraphrases, which is more difficult to assess by global similarity measures such as the measure of Lin and Pantel.

## 2 Background and Related Work

Recently, several works addressed the task of automatically acquiring paraphrase patterns from corpora. (Barzilay & McKeown 01) use sentence alignment to identify paraphrases from a corpus of multiple English translations of the same text. In another more recent work, (Pang et al. 03) also use a parallel corpus of Chinese-English translations to build finite state automata for paraphrase patterns, based on syntactic alignment of corresponding sentences.

(Shinyama et al. 02) learn structural paraphrase templates for Information extraction from a comparable corpus of news articles from different news sources over a common period of time. Similar news article pairs from

| subject | secretary_general_boutros_boutros_ghali |
|---------|------------------------------------------|
| object  | implementation_of_deal |
| modifier| after |

(A) verb: delay

| subject | iraqi_force |
|---------|-------------|
| object  | kurdish_rebel |
| pp-on   | august_31 |

(B) verb: attack

Figure 1: extracted verb instances for sentence "But U.N. Secretary-General Boutros Boutros-Ghali delayed implementation of the deal after Iraqi forces attacked Kurdish rebels on August 31."

the different news sources are identified based on document similarity. Sentence pairs (from a given pair of similar articles) are then identified based on the similarity of Named Entities in the matching sentences. (Barzilay and Lee 03) also utilizes a comparable corpus of news articles to learn paraphrase patterns, which are represented by word lattice pairs. Patterns originating from the same day but from different newswire agencies are matched based on entity overlap.

We compare our results to those of the algorithm by (Lin & Pantel 01), which extracts paraphrase-like inference rules for question answering from a single source corpus. The underlying assumption in their work is that paths in dependency trees that connect similar syntactic arguments (slots) are close in meaning. Rather then considering a single feature vector that originates from the arguments in both slots, vector-based similarity was computed separately for each slot, using the similarity measure of (Lin 98). The similarity of a pair of binary paths was defined as the geometric mean of the similarity values that were computed for each of the two slots.

## 3 Algorithm

Our proposed algorithm identifies candidates of corresponding verb paraphrases within pairs of sentences. We define a *verb instance pair* as a pair of occurrences of two distinct verbs in the corpus. A *verb type pair* is a pair of verbs detected as a candidate lexical paraphrase.

### 3.1 Preprocessing and Representation

Our algorithm relies on a syntactic parser to identify the syntactic structure of the corpus sentences, and to identify verb instances. We treat the corpus uniformly as a set of distinct sentences, regardless of the document or paragraph they belong to. For each verb instance we extract the various syntactic components that are related directly to the verb in the parse tree. For each such component we extract its lemmatized head, which is possibly extended to capture a semantically specified constituent. We extended the heads with any lexical modifiers that constitute a multi-word term, noun-noun modifiers, numbers and prepositional 'of' complements.

Verb instances are represented by the vector of syntactic modifiers and their lemmatized fillers. For illustration, Figure 1 shows an example sentence and the vector representations for its two verb instances.

### 3.2 Identifying candidate verb instance pairs (filtering)

We apply various filters in order to verify that two verb instances are likely to be paraphrases describing the same event. This is an essential part of the algorithm since we do not rely on the high a-priori likelihood for finding paraphrases in matching parts of comparable corpora.

We first limit our scope to pairs of verb instances that share a common (extended) subject and object which are not pronouns. Otherwise, if either the subject or object differ between the two verbs then they are not likely to refer to the same event in a manner that allows substituting one verb with the other.

Additionally, we are interested in identifying sentence pairs with a significant overall term overlap, which further increases paraphrase likelihood for the same event. This is achieved with a standard (Information Retrieval style) vector-based approach, with tf-idf term weighting (Frakes and Baeza-Yates 92)

- $tf(w)$ = freq($w$) in sentence
- $idf(w) = \log(N / \text{freq}(w)$ in corpus)
  where $N$ is the total number of tokens in the corpus.

Sentence overlap is measured simply as the dot product of the two vectors. We intentionally disregard any normalization factor (such as in the Cosine measure) in order to assess the "absolute" degree of overlap, while allowing longer sentences to include also non-matching parts that might correspond to complementary aspects of the same event. Verb instance pairs whose sentence overlap is below a specified threshold are filtered out.

An additional assumption is that events have a unique propositional representation and hence verb instances with contradicting vectors are not likely to describe the same event. We therefore filter verb instance pairs with contradicting propositional information – a common syntactic relation with different arguments. As an example, the sentence "Iraqi forces captured Kurdish rebels on August 29." Has a contradicting 'on' preposition argument with the sentence from Figure 1B ("August 29" vs. "August 31").

### 3.3 Computing paraphrase score of verb instance pairs

Given a verb instance pair (after filtering), we want to estimate the likelihood that the two verb instances are paraphrases of the same fact or event. We thus assign a paraphrase likelihood score for a given verb instance pair $I_{v1,v2}$, which corresponds to instances of the verb types $v_1$ and $v_2$ with overlapping syntactic components $p_1, p_2, \ldots p_n$. The score corresponds (inversely) to the estimated probability that such overlap had occurred by chance in the entire corpus, capturing the view that a low overlap probability (i.e. low probability that the overlap is due to chance) correlates with paraphrase likelihood. We estimate the overlap probability by assuming independence of the verb and each of its syntactic components as follows:

$$(1) \quad P(I_{v1,v2}) = P(overlap) = P(v_1, p_1 \ldots p_n) P(v_2, p_1 \ldots p_n)$$
$$= P(v_1) P(v_2) \prod_{i=1}^{n} P(p_i)^2$$

Where the probabilities were calculated using Maximum Likelihood estimates based on the verb and argument frequencies in the corpus.

### 3.4 Computing paraphrase score for verb type pairs

When computing the score for a verb type pair we would like to accumulate the evidence from its corresponding verb instance pairs. Following the vein of the previous section we try to estimate the joint probability that these different instance pairs occurred by chance. Assuming instance independence, we would like to multiply the overlap probabilities obtained for all instances. We have found, though, that verb instance pairs whose two verbs share the same subject and object are far from being independent (there is a higher likelihood to obtain additional instances with the same subject-object combination). To avoid complex modeling of such dependencies we picked only one verb instance pair for each subject-object combination, taking the one with lowest probability (highest score). This yields the set $T(v_1,v_2)=(I_1...I_n)$ of best scoring (lowest probability) instances for each distinct subject and object components. Assuming independence of occurrence probability of these instances, we estimate the probability $P(T(v_1,v_2))=\Pi P(I_i)$, where $P(I)$ is calculated by Equation (1) above. The score of a verb type pair is given by:

$$(2) \quad score(v_1, v_2) = -\log P(T(v_1, v_2))$$

## 4 Evaluation and Analysis

### 4.1 Setting

We ran our experiments on the first 15-million word (token) subset of the *Reuters Corpus.*[1] The corpus sentences were parsed using the Minipar[2] dependency parser (Lin 93). 6,120 verb instance pairs passed filtering (with overlap threshold set to 100). These verb instance pairs derive 646 distinct verb type pairs, which were proposed as candidate lexical paraphrases along with their corresponding paraphrase score.

---

[1] Known as *Reuters Corpus, Volume 1, English Language, 1996-08-20 to 1997-08-19*, provided by Reuters on CD.
[2] http://www.cs.umanitoba.ca/~lindek/minipar.htm

| 1- | <fall, rise> | 6+ | <drop, fall> | 62+ | <honor, honour> | 362+ | <bring, take> |
|---|---|---|---|---|---|---|---|
| 2+ | <close, end> | 7+ | <regard, view> | 122+ | <advance, rise> | 422+ | <note, say> |
| 3+ | <post, report> | 8+ | <cut, lower> | 182+ | <benefit, bolster> | 482- | <export, load> |
| 4+ | <recognize, recognize> | 9- | <rise, shed> | 242+ | <approve, authorize> | 542+ | <downgrade, relax> |
| 5+ | <fire, launch> | 10+ | <fall, slip> | 302+ | <kill, slaughter> | 602+ | <create, establish> |

Table 2: Example of system output with judgments.

The correctness of the extracted verb type pairs was evaluated over a sample of 215 pairs (one third of the complete set) by two human judges, where each judge evaluated one half of the sample. In a similar vein to related work in this area, judges were instructed to evaluate a verb type pair as a *correct* paraphrase only if the following condition holds: one of the two verbs can replace the other within some sentences such that the meaning of the resulting sentence will entail the meaning of the original one. Otherwise the pair is judged as *incorrect*.

Notice that the judgment criterion requires that the lexical paraphrase would hold in some contexts, but not necessarily all. To assist the judges in assessing a given verb type pair they were presented with example sentences from the corpus that include some matching contexts for the two verbs (e.g., sentences in which both verbs have the same subject or object). Notice that the judgment criterion allows for "directional" paraphrases, such as <invade, enter> or <slaughter, kill>, where the meaning of one verb entails the meaning of the other, but not vice versa. Additionally, a verb type pair was judged as correct if a verb is part of a multi-word expression, which by itself is a paraphrase of the other verb. For example, the pair <reject, turn> is judged as correct, since the expression "turn down" is a correct paraphrase of "reject". The judges based their decision in such cases on the multi-word expressions found in the example sentences they were viewing. The motivation for judging such verb type pairs as correct is that they represent cases in which our current algorithm performed correctly and identified verb instance pairs that are indeed paraphrases of each other; identification of complex multi-word expressions was beyond the scope of the current study.

### 4.2 Results of the paraphrase identification algorithm

Figure 2 shows the precision vs. recall results for each judge over the given test-sets. The evaluation was conducted separately also by the authors on the full set of 646 verb pairs, obtaining comparable results to the independent evaluators. In terms of agreement, the Kappa value (measuring pair wise agreement discounting chance occurrences) between the authors and the independent evaluators' judgments were 0.61 and 0.63, which correspond to a *substantial* agreement level (Landis & Koch 77).

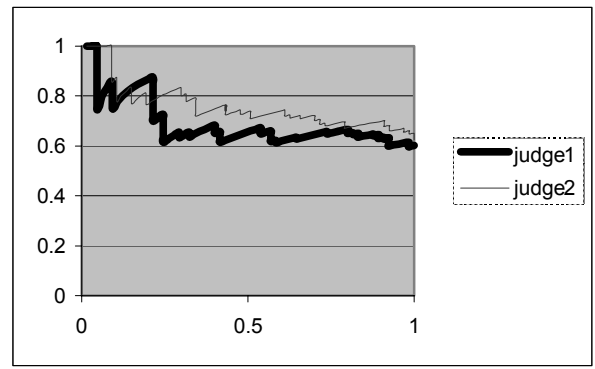

Figure 2: precision (*y* axis) recall (*x* axis) curves of system paraphrases by judge (verb type pairs sorted by system score).

The overall precision for the complete test sample is 61.4% accuracy, with a confidence interval of [56.1,66.7] at the 0.05 significance level.

Table 2 shows the top 10 lexical paraphrases, and a sample of the remaining ones, achieved by our system along with the annotators' judgments. Table 3 shows "correct" sentence pairs describing a common event, which were identified by our system as candidate paraphrase instances.

| Ieng Sary on Wednesday formally **announced** his split with top Khmer Rouge leader Pol Pot, and said he had formed a rival group called the Democratic National United Movement. | In his Wednesday announcement Ieng Sary, who was sentenced to death in absentie for his role in the Khmer Rouge's bloody rule, **confirmed** his split with paramount leader Pol Pot. |
|---|---|
| Campbell is **buying** Erasco from Grand Metropolitan Plc of Britain for about $210 million. | Campbell is **purchasing** Erasco from Grand Metropolitan for approximately US$210 million. |
| The stock of Kellogg Co. **dropped** Thursday after the giant cereal maker warned that its earnings for the third quarter will be 20 percent below a year ago. | The stock of Kellogg Co. **fell** Thursday after it warned about lower earnings this year and sparked concerns that it could resort to increased promotions to protect its leading market share, analysts said. |
| Slovenian President Milan Kucan **opened** a second round of consultations with political parties on Monday to try to agree on a date for a general election which must take place between October 27 and December 8. | - Slovenian President Milan Kucan on Monday **started** a second round of consultations with political parties concerning the election date. |

Table 3: Examples of "correct" paraphrase instance pairs

| Last Friday, the United States **announced** punitive charges against China's 1996 textile and apparel quotas, citing transhipments of Chinese textiles in voilation of a 1994 trade agreement | China on Saturday urged the United States to **rescind** punitive charges against Beijing's 1996 textile and apparel quotas and threatened retaliatory action. |
|---|---|
| Rand Financials notably **bought** October late while Chicago Corp and locals lifted December into by stops. | Rand Financials notably **sold** October late while locals pressured December. |
| Municipal bond yields **dropped** as much as 15 basis points in the week ended Thursday, erasing increases from the week before. | Municipal bond yields **jumped** as much as 15 basis points over the week ended Thursday on top of similar increases the week before. |
| French shares **opened** lower, ignoring gains on Wall Street and other European markets, due to renewed pressure on the franc and growing worries about possible strike action in the autumn, dealers said. | French shares **closed** sharply lower on Wednesday due to a weaker franc amid evaporating hopes of a German rate cut on Thursday, but the market managed to remain above the 2,000 level and did keep some of Tuesday's gains. |

Table 4: Examples of "misleading" instance pairs

An analysis of the incorrect paraphrases showed that roughly one third of the errors captured verbs with contradicting semantics or antonyms (e.g. <rise, fall>, <buy, sell>, <capture, evacuate>) and another third were verbs that tend to represent correlated events with strong semantic similarity (e.g. <warn, attack>, <reject, criticize>). These cases are indeed quite difficult to distinguish from true paraphrases since they tend to occur in a corpus with similar overlapping syntactic components and within quite similar sentences. Table 4 shows example sentences demonstrating the difficulties posed by such cases, which turn to be quite similar in their nature to anecdotal paraphrases of the same event that might be spread along a single corpus (cf. Table 3). Systems tailored for comparable corpora are less likely to confront such problematic sentence pairs for they are less likely to occur in corresponding stories.

It should be noticed that our evaluation was performed at the verb *type* level. We have not evaluated directly the "correctness" of the individual paraphrase instance pairs extracted by our method (i.e. whether the two instances in a paraphrase pair indeed refer to the same fact). Such evaluation is planned for future work.

Finally, a general problematic (and rarely addressed) issue in this area of research is how to evaluate the coverage or recall of the extraction method relative to a given corpus.

### 4.3 Comparison with (Lin & Pantel 01)

We applied the algorithm of (Lin & Pantel 01), denoted here as the LP algorithm, and computed their similarity score for each pair of verb types in the corpus. To implement the method for lexical verb paraphrases, each verb type was considered as a distinct path whose subject and object play the roles of the *X* and *Y* slots (cf. Section 2).

| 1 | 0.62 | misread, misjudge | 6 | 0.23 | mark_down, decontrol | 11 | 0.20 | flatten, steepen | 16 | 0.18 | trumpet, drive_home |
| --- | --- | --- | --- | --- | --- | --- | --- | --- | --- | --- | --- |
| 2 | 0.29 | barricade, sandbag | 7 | 0.22 | subsidize, subsidise | 12 | 0.20 | mainline, pip | 17 | 0.17 | marshal, beleaguer |
| 3 | 0.27 | disgust, mystify | 8 | 0.21 | wake_up, divine | 13 | 0.20 | misinterpret, relive | 18 | 0.17 | dwell_on, feed_on |
| 4 | 0.27 | jack, decontrol | 9 | 0.21 | thrill, personify | 14 | 0.19 | remarry, flaunt | 19 | 0.16 | scrutinize, misinterpret |
| 5 | 0.25 | Pollinate, pod | 10 | 0.20 | mark_up, decontrol | 15 | 0.18 | distance, dissociate | 20 | 0.16 | disable, counsel |

Table 5: top 20 verb pairs from similarity system.

As it turned out, the similarity score of LP does not behave uniformly across all verbs. For example, many of the top 20 highest scoring verb pairs are quite erroneous (see Table 5), and do not constitute lexical paraphrases (compare with the top scoring verb pairs for our system in Table 2). The similarity scores do seem meaningful within the context of a single verb *v*, such that when sorting all other verbs by the LP score of their similarity to *v* correct paraphrases are more likely to occur in the upper part of the list. Yet, we are not aware of a criterion that predicts whether a certain verb has few good paraphrases, many or none. Given this behavior of the LP score we chose the following procedure to create a test sample for the LP algorithm that is comparable to our own test sample. For each verb type pair ($v_1$,$v_2$) in our sample we chose randomly one of the two verbs as a "pivot" (assume the pivot is $v_1$). We then identified the rank *k* of $v_2$ among all verb type pairs that include $v_1$, when sorted by the paraphrase score of our method. That is, $v_2$ is the *k*'th most likely paraphrase for $v_1$ according to our method. Finally, we took the *k*'th verb in the LP similarity list of $v_1$, say $v_j$, and inserted the pair ($v_1$,$v_j$) to the LP test sample. Thus, both test set samples contain verb type pairs with equivalent similarity rankings relative to the other sample. Notice that this procedure is favorable to the LP method for it is evaluated at points (verb and rank) that where predicted by our method to correspond to a likely paraphrase.

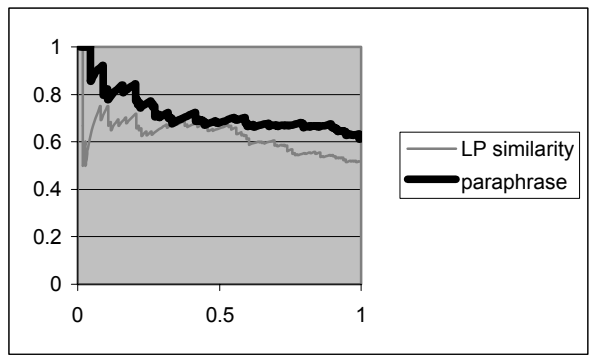

Figure 3: Precision recall curve for our paraphrase method and LP similarity.

The resulting 215 verb pairs were evaluated by the judges along with the sample for our method, while the judges did not know which system generated each pair. The overall precision on the LP method for the sample was 51.6%, with a confidence interval of [46.1,57.1] at the 0.05 significance level[3]. The LP results for this sample were thus about 10 points lower than the results for our comparable sample, but the two confidence intervals overlap slightly. It is interesting to note that the precision of the LP algorithm over all pairs of rank 1 was also 51%, demonstrating that just rank on its own is not a good basis for paraphrase likelihood.

Figure 3 shows overall recall vs. precision from both judges for the two systems. The results above show that the precision of the vector-based LP method may be regarded as comparable to our instance-based method, in cases where one of the two verbs was identified by our method to have a corresponding number of paraphrases. The obtained level of accuracy for these cases is substantially higher than for the top scoring pairs by LP. This suggests that our approach can be combined with the vector-based approach to obtain higher reliability for verb pairs that were extracted from actual paraphrase instances.

---

[3] We noticed that the LP method, which computes similarity scores separately for subjects and objects data, seems to perform better at identifying paraphrases than the classic vector-based similarity approach (Lin 98, Dagan 2000), which computes similarity once based on all features together.

As an example, the top four verbs similar to 'buy' based on the LP algorithm (along with their score) are: sell (0.069), purchase (0.052), acquire (0.039) and import (0.035). However, the top four verbs based on our method are: purchase (171), acquire (102), sell (100) and take (91). This representative example demonstrates that although the output of the two systems is similar, many instances that contribute to the high similarity of antonymous pairs such as <buy, sell> are filtered out by our system and obtain an overall lower rank then true synonyms.

## 5 Conclusion and Future work

This paper presented an algorithm for extracting lexical verb paraphrases from a single corpus. To the best of our knowledge, this is the first attempt to identify actual paraphrase instances in a single corpus and to extract paraphrase patterns directly from them. The evaluation suggests that such an approach is indeed viable, based on algorithms that are geared to overcome many of the "misleading" cases that are typical for a single corpus (in comparison to comparable corpora). Furthermore, a preliminary comparison suggests that an instance-based approach may be combined with a vector-based approach in order to assess better the paraphrase likelihood for many verb pairs. Future research is planned to extend the approach to handle more complex paraphrase structures and to increase its performance by relying on additional sources of evidence.

## Acknowledgements
We would like to thank Yuval Berger, Alfred Frohlich and Maayan Geffet for their help in evaluation and Moshe Koppel and Zvika Marx for helpful discussions.